\documentclass[runningheads]{llncs}

\usepackage{amsfonts}
\usepackage{amssymb}
\usepackage{amsmath}
\usepackage[latin1]{inputenc}
\usepackage[dvips]{graphicx}

\renewenvironment{proof}{\list{}{\parsep 0em \leftmargin 2em \rightmargin 0em}%
                \item\relax Proof: }{\hfill $\square$ \endlist}

\DeclareMathOperator{\comp}{\theta}
\DeclareMathOperator{\SUM}{SUM}
\DeclareMathOperator{\MIN}{MIN}
\DeclareMathOperator{\MAX}{MAX}
\DeclareMathOperator{\AVG}{AVG}

\newcommand{\LL}{\ensuremath{\mathcal{L}}}
\newcommand{\OO}{\ensuremath{\mathcal{O}}}
\renewcommand{\AA}{\ensuremath{\mathcal{A}}}
\newcommand{\db}{\ensuremath{db}}

\newcommand{\freq}{\ensuremath{\mathcal{F}}}

\newcommand{\txtrm}[1]{\text{\upshape\sffamily #1}}
\newcommand{\CC}{\ensuremath{\mathcal{C}}}
\newcommand{\tCC}{{^t\CC}}

\newcommand{\cl}{\txtrm{cl}}

\newcommand{\Cfreq}{\CC_{\gamma\txtrm{-freq}}}

\newcommand{\Cclos}{\CC_{\txtrm{close}}}

\newcommand{\et}{\wedge}
\newcommand{\ou}{\vee}
\newcommand{\equ}{\Leftrightarrow}

\newcommand{\set}[1]{\left\{ #1 \right\}}
\newcommand{\size}[1]{\left| #1 \right|}

\newcommand{\sseq}{\subseteq}

\newcommand{\ra}{\,r_a\,}
\newcommand{\ro}{\,r_o\,}

\newcommand{\at}[1]{a_#1}
\newcommand{\ob}[1]{o_#1}

\newcommand{\defnt}[1]{\textbf{#1}}

\author{Baptiste Jeudy\inst{1} and Fran\c{c}ois Rioult\inst{2}}

\institute{Équipe Universitaire de Recherche en Informatique de St-Etienne,\\
  Université de St-Etienne, France. \\
  \email{baptiste.jeudy@univ-st-etienne.fr}
  \and 
GREYC - CNRS UMR 6072,\\
Université de Caen Basse-Normandie, France.\\
\email{francois.rioult@info.unicaen.fr}}

\title{\textbf{Database Transposition for\\ Constrained (Closed) Pattern Mining}}

\begin{document}

\maketitle
\begin{abstract}

Recently, different works proposed a new way to mine patterns in databases with pathological
size. For example, experiments in genome biology usually provide databases with thousands of
attributes (genes) but only tens of objects (experiments). In this case, mining the ``transposed''
database runs through a smaller search space, and the Galois connection allows to infer the closed
patterns of the original database.

We focus here on constrained pattern mining for those unusual databases and give a theoretical
framework for database and constraint transposition. We discuss the properties of constraint
transposition and look into classical constraints. We then address the problem of generating the
closed patterns of the original database satisfying the constraint, starting from those mined in the
``transposed'' database. Finally, we show how to generate all the
patterns satisfying the constraint from the closed ones.

\end{abstract}

\section{Introduction}

Frequent pattern mining is now well mastered, but these patterns, like association rules, reveal to
be too numerous for the experts and very expensive to compute. They have to be filtered or
constrained.  However, mining and constraining have to be done jointly (pushing the constraint) in
order to avoid combinatorial explosion~\cite{jeudyboulicautida}. Mining under complex constraint has
become today a hot topic and the subject of numerous works
(e.g.,~\cite{jeudyboulicautida,boulicautjeudy01a,ngetal98,peietal01,deraedtkramer01,bgk02}).
Moreover, new domains are interested in our applications, and data schemes vary consequently.  In
genome biology, biological experiments are very expensive and time consuming. Therefore, only a
small number of these experiments can be processed. However, thanks to new devices (such as
biochips), experiments can provide the measurements of the activity of thousands of genes. This
leads to databases with lots of columns (the genes) and few rows (the experiments).

Numerous works present efficient algorithms which mine the patterns satisfying a user defined
constraint in large databases. This constraint can combine minimum and maximum frequency threshold
together with other syntactical constraints.  These algorithms are designed for databases with up to
several millions of rows.  However, their complexity is exponential in the number of columns and thus
they are not suited for databases with too many columns, like those encountered in genome biology.

Recently, two propositions were done to solve this problem: instead of mining the original database,
these algorithms work on the ``transposed'' database, i.e., columns of the original database become
rows in the ``transposed'' database and rows becomes columns (this is indeed the same database but
with a different representation). Therefore the ``transposed'' database has significantly less columns
than the original one.  The CARPENTER algorithm \cite{pct03} is specifically designed for mining the
frequent closed patterns, and our proposition \cite{rbc03,rc04} uses a classical algorithm for
mining closed patterns with a monotonic (or anti-monotonic) constraint. Both approaches use the
transposition principle, however the problem of mining under constraints is not fully studied,
specially for complex constraints (i.e., conjunction and disjunction of simple constraints).

In this paper, we study this problem from a theoretical point of view.  Our aim is to use classical
algorithms (constrained pattern mining algorithms or closed patterns mining algorithms) in the
``transposed'' database and to use their output to regenerate patterns of the original database instead
of directly mining in the original database.

\noindent There are several interesting questions which we will therefore try to answer:
\begin{enumerate}
\item What kind of information can be gathered in the
``transposed'' database on the patterns of the original database ?
\item Is it possible to ``transpose'' the constraints ? I.e., given
a database and a constraint, is it possible to find a ``transposed'' constraint such that 
mining the ``transposed'' database with the ``transposed'' constraint gives information
about the patterns which satisfy the original constraint in the original database ? 
\item How can we regenerate the closed patterns in the original database from the
patterns extracted in the ``transposed'' database ? 
\item How can we generate {\em all} the itemsets satisfying a constraint using 
  the extracted closed patterns.
\end{enumerate}
These questions will be addressed respectively in Sec.~\ref{sec:definitions}, 
\ref{sec:tconst}, \ref{sec:closed-it}~and~\ref{sec:itemset-mining}.

The organization of the paper is as follows: we start Sec.~\ref{sec:definitions} by recalling some
usual definitions related to pattern mining and Galois connection. Then we show in
Sec.~\ref{sec:tconst} how to transpose usual and complex constraints. Section~\ref{sec:closed-it} is
a complete discussion about mining constrained closed patterns using the ``transposed'' database and
in Sec.~\ref{sec:itemset-mining}
we show how to use this to compute all (i.e., not only closed)
the patterns satisfying a constraint.
Finally Sec.~\ref{sec:conclusion} is a short conclusion.

\section{Definitions}
\label{sec:definitions}
To avoid confusion between rows (or columns) of the original database and rows (columns) of the
``transposed'' database, we define a database as a relation between two sets : a set of attributes and
a set of objects.  The set of \defnt{attributes} (or items) is denoted $\AA$ and the set of
\defnt{objects} is $\OO$.  The \defnt{attribute space} $2^\AA$ is the collection of the subsets of $\AA$
and the \defnt{object space} $2^\OO$ is the collection of the subsets of $\OO$.  An \defnt{attribute
  set} (or \defnt{itemset} or attribute pattern) is a subset of $\AA$. An \defnt{object set} (or
\defnt{object pattern}) is a subset of $\OO$. A \defnt{database} is a subset of $\AA \times \OO$.

In this paper we consider that the database has more attributes than objects and that we are
interested in mining attributes sets.  The database can be represented as an adjacency matrix where
objects are rows and attributes are columns (original representation) or where objects are
columns and attributes are rows (transposed representation).

\begin{table}
  \caption{Original and transposed representations of a database. The attributes are $\AA =
  \set{\at{1},\at{2},\at{3},\at{4}}$ and the objects are $\OO=\set{\ob{1},\ob{2},\ob{3}}$. We use a
  string notation for object sets or itemsets, e.g., $\at{1}\at{3}\at{4}$ denotes the itemset
  $\set{\at{1},\at{3},\at{4}}$ and $\ob{2}\ob{3}$ denotes the object set $\set{\ob{2},\ob{3}}$. 
  This dataset is used in all the examples. } 

  \hfill
  \begin{tabular}{|c|c|}
    \hline
    ~object~ &~attribute pattern~\\
    \hline
    $\ob{1}$ & $\at{1}\at{2}\at{3}$\\
    $\ob{2}$ & $\at{1}\at{2}\at{3}$\\
    $\ob{3}$ & $\at{2}\at{3}\at{4}$\\
    \hline
  \end{tabular}
  \hfill
  \begin{tabular}{|c|c|}
    \hline
    ~attribute~ & ~object pattern~\\
    \hline
    $\at{1}$ & $\ob{1}\ob{2}$\\
    $\at{2$} & $\ob{1}\ob{2}\ob{3}$\\
    $\at{3}$ & $\ob{1}\ob{2}\ob{3}$\\
    $\at{4$} & $\ob{3}$\\
    \hline
  \end{tabular}
  \hspace*{\fill}
  \label{tab:database}
\end{table}

\subsection{Constraints}
Given a database, a \defnt{constraint} $\CC$ on an attribute set (resp. object set) is a boolean
function on $2^\AA$ (resp. on $2^\OO$). Many constraints have been used in previous works. One of
the most popular is the minimum frequency constraint which requires an itemset to be present in more
than a fixed number of objects. But we can also be interested in the opposite, i.e., the maximum
frequency constraint. Other constraints are related to Galois connection (see
Sect.~\ref{sec:galois}), such as closed~\cite{bastideetal00b} patterns, free~\cite{bbr03},
contextual free~\cite{boulicautjeudy01a} or key~\cite{bastideetal00b} patterns, or even
non-derivable~\cite{caldersetal02} or emergent~\cite{scr042,dl99} patterns. There are also
syntactical constraints, when one focuses only on itemsets containing a fixed pattern (superset
constraint), contained in a fixed pattern (subset constraint), etc. Finally, when a numerical value
(such as a price) is associated to items, aggregate functions such as sum, average, min, max, etc.
can be used in constraints~\cite{ngetal98}.

A constraint $\CC$ is \defnt{anti-monotonic} if $\forall A,B \; (A \sseq B \et \CC(B)) \Longrightarrow
\CC(A)$.  A constraint $\CC$ is \defnt{monotonic} if $\forall A,B\; (A \sseq B \et \CC(A))
\Longrightarrow \CC(B)$.  In both definitions, $A$ and $B$ can be attribute sets or object sets. The
frequency constraint is anti-monotonic, like the subset
constraint. The anti-monotonicity property is
important, because level-wise mining algorithms most of time use it to prune the search space.
Indeed, when a pattern does not satisfy the constraint, its specialization neither and can be
pruned~\cite{agrawaletal96}.  

Simple composition of constraints has good properties: the conjunction or the disjunction of two
anti-monotonic (resp. monotonic) constraints is anti-monotonic (resp. monotonic). The negation of an
anti-monotonic (resp. monotonic) constraints is monotonic (resp. anti-monotonic).

\subsection{Galois Connection}
\label{sec:galois}

The main idea underlying our work is to use the strong connection between the itemset lattice
$2^\AA$ and the object lattice $2^\OO$ called the \defnt{Galois connection}. This connection was first
used in pattern mining when closed itemset mining algorithms were proposed~\cite{pasquieretal99},
while it relates to many works in concept learning~\cite{mn00,WIL92}.

\noindent Given a database $\db$, the Galois operators $f$ and $g$ are defined as:
\begin{itemize}
\item $f$, called {\em intension}, is a function from $2^\OO$ to $2^\AA$ defined by
  $$f(O) = \set{a \in \AA \mid \forall o \in O,\; (a,o) \in \db },$$
\item $g$, called {\em extension}, is a function from $2^\AA$ to $2^\OO$ defined by
  $$g(A) = \set{o \in \OO \mid \forall a \in A,\; (a,o) \in \db }.$$
\end{itemize}

Given an itemset $A$, $g(A)$ is also called the {\em support set} of $A$ in $\db$.  It is also the
set of objects for which all the attributes of $A$ are true.  The \defnt{frequency} of $A$ is
$\size{g(A)}$ and is denoted $\freq(A)$.

Both functions enable us to link the attribute space to the object space. However, since both spaces
have not the same cardinality, there is no one to one mapping between them\footnote{This is
  fortunate since the whole point of transposition is to explore a smaller space.}. This means that
several itemsets can have the same image in the object space and conversely. We thus define two
equivalence relations $\ra$ and $\ro$ on $2^\OO$ and $2^\AA$:
\begin{itemize}
\item if $A$ and $B$ are two itemsets, $A \ra B$ if $g(A)=g(B)$,
\item if $O$ and $P$ are two sets of objects, $O \ro P$ if $f(O)=f(P)$.
\end{itemize}

In every equivalence class, there is a particular element: the largest (for inclusion) element of an
equivalence class is unique and is called a \defnt{closed attribute set} (for $\ra$) or a \defnt{closed
  object set} (for $\ro$).

The Galois operators $f$ and $g$ lead by composition to two \defnt{closure} operators, namely $h = f
\circ g$ and $h' = g \circ f$. They relate to lattice or hypergraph theory and have good
properties~\cite{SS02}.  The closed sets are then the fixed points of the closure operators and the
closure of a set is the closed set of its equivalence class.  In the following we will indifferently
refer to $h$ and $h'$
with the notation $\cl$. We denote $\Cclos$ the constraint which is satisfied by the itemsets
or the object sets which are closed.

If two itemsets are equivalent, their images are equal in the object space.  There is therefore no
mean to distinguish between them if the mining of the closed patterns is performed in the object space. So, by using
the Galois connection to perform the search in the object space instead of the attribute space, we
will gather information about the equivalence classes of $\ra$ (identified by their closed pattern),
not about all individual itemsets. This answers the first question of the introduction, i.e. what kind
of information can be gathered in the transposed database on the patterns of the original database.
At best, we will only be able to discover closed patterns.

\begin{figure}
  \centering \includegraphics[scale = 1]{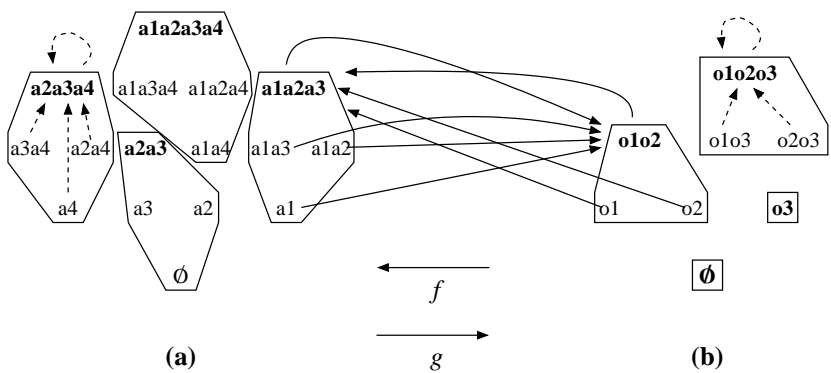}
  \caption{The equivalence classes for $\ra$ in the itemset lattice (a) and for $\ro$ in the object 
    set lattice (b) built on the database of Tab.~\ref{tab:database}.  The closed sets are in bold
    face. The arrows represent the $f$ and $g$ operators between the $\at{1}\at{2}\at{3}$ and
    $\ob{1}\ob{2}$ equivalence classes. The dotted arrows represent the closure operators $h$ and
    $h'$ }
  \label{fig:treillis}
\end{figure}

\begin{property} \label{prop:decr} Some properties of $f$ and $g$.
  \begin{itemize}
  \item $f$ and $g$ are decreasing w.r.t. the inclusion order: if $A \sseq B$ then $g(B) \sseq g(A)$
    (resp. $f(B) \sseq f(A)$)
  \item If $A$ is an itemset and $O$ an object set, then $g(A)$ is a closed object set and $f(O)$ a
    closed itemset
  \item fixed point: $A$ is closed if and only if $f(g(A)) = \cl(A) = A$ (resp. $g(f(O)) = \cl(O) =
    O$)
  \item $f \circ g \circ f =f$ and $g \circ f \circ g =g$
  \item $A \sseq \cl(A)$
  \end{itemize}
\end{property}

In the Galois connection framework, the association of a closed pattern of attributes and the
corresponding closed pattern of objects is called a {\em concept}. Concept
learning~\cite{mn00,WIL92} has led to classification tasks and clustering processes. We use this
connection in this article through the link it provides between the search spaces $2^\AA$ and
$2^\OO$.

\begin{example}
  In Fig.~\ref{fig:treillis}, the closed objects sets are $\emptyset$, $\ob{3}$, $\ob{1}\ob{2}$, and
  $\ob{1}\ob{2}\ob{3}$. The closed itemsets are $\at{2}\at{3}$, $\at{2}\at{3}\at{4}$,
  $\at{1}\at{2}\at{3}$ and $\at{1}\at{2}\at{3}\at{4}$. Since $g(\ob{1}\ob{2})=\at{1}\at{2}\at{3}$
  and $f(\at{1}\at{2}\at{3})=\ob{1}\ob{2}$, $(\at{1}\at{2}\at{3},\ob{1}\ob{2})$ is a concept. The
  others are $(\at{2}\at{3},\ob{1}\ob{2}\ob{3})$, $(\at{2}\at{3}\at{4},\ob{3})$,
  $(\at{1}\at{2}\at{3}\at{4},\emptyset)$.
\end{example}

Closed sets of attributes are very useful for algorithms with support constraint, because they
share, as maximal element of the equivalence class $\ra$, the same frequency with all patterns in
the class. Closed set mining is now well known~\cite{fmn03}, and
frequent closed patterns are known to be less numerous than frequent patterns~\cite{bgkm02,caldersetal02}.
Today's approaches relate to closed sets with constraints mining~\cite{brb04}. These patterns 
are good candidates for constituting relevant concepts, which associate at the same 
time the attributes and the objects. For example, biologists want to constraint their search 
to attribute patterns containing some specific genes, with a specified
maximum length. They also will be interested in analyzing the other
part of the concept. We specifically
address here the problem of constrained closed mining in databases with more attributes than
objects.

\section{Constraint Transposition}
\label{sec:tconst}

Most algorithms extracting closed patterns are search algorithms. The size of the search space
strongly determines their performance~\cite{fmn03}.  In our context, the object space $2^\OO$ is smaller
than the attribute space $2^\AA$.  We therefore choose to search the closed patterns in the smaller
space ($2^\OO$) by transposing the database. In order to mine under constraint, we study in this
section how we can adapt constraints to the new transposed database, i.e., how we can transpose 
constraints. We will therefore answer question~2 of the introduction.

\subsection{Definition and Properties}

Given an itemset constraint $\CC$, we want to extract the collection $I$ of itemsets, $I = \set{ A
  \sseq \AA \mid \CC(A)}$. Therefore, we want to find in the transposed database a collection $T$ of
object sets (if it exists) such that the image by $f$ of this collection is $I$, i.e., $\set{f(O)
  \mid O \in T}= I$.  Since $f(O)$ is always a closed itemset, this is only possible if the
collection $I$ contains only closed itemsets (i.e., if the constraint $\CC$ includes the $\Cclos$
constraint).  In this case, a solution for $T$ is the collection $\set{O \sseq \OO \mid \CC(f(O))}$
which leads to the following definition of a transposed constraint:

\begin{definition}[Transposed constraint]\label{def:tc}
  Given a constraint $\CC$, we define the transposed constraint $\tCC$ on a closed pattern $O$ of
  objects as:
  $$ \tCC(O) = \CC(f(O)).$$
\end{definition}

\begin{example}
  Consider the itemset constraint $\CC(A)=(\at{1}\in A)$. Its transposed constraint is (by definition)
  $\tCC(O)=(\at{1}\in f(O))$. Using the dataset of Tab.~\ref{tab:database}, the object sets that
  satisfy $\tCC$ are
  $T=\set{\ob{1},\ob{2},\ob{1}\ob{2},\ob{1}\ob{3},\ob{2}\ob{3},\ob{1}\ob{2}\ob{3}}$. If we compute
  $\set{f(O) \mid O \in T}$, we get $\set{\at{1}\at{2}\at{3},\at{1}\at{2}\at{3}\at{4}}$ which are
  exactly the closed itemsets that satisfy $\CC$. Theorem~\ref{th:1} will show that this
  is always the case.
\end{example}

It is interesting to study the effect of transposition w.r.t. the monotonicity or anti-monotonicity
of constraints, since many mining algorithms rely on them for efficient pruning:

\begin{proposition}
  If a constraint $\CC$ is monotonic (resp. anti-monotonic), the transposed constraint
  $\tCC$ is anti-monotonic (resp. monotonic).
\end{proposition}
\begin{proof}
$f$ and $g$ are decreasing (Prop.~\ref{prop:decr}), which inverts monotonicity and
anti-monotonicity.  
\end{proof}

Since we also want to deal with complex constraints (i.e., constraints built with elementary
constraints using boolean operators), we need the following:

\begin{proposition}\label{prop:bool_op}
  If $\CC$ and $\CC'$ are two constraints then:
  $$ ^t(\CC \et \CC') = \tCC \et \tCC'$$
  $$ ^t(\CC \ou \CC') = \tCC \ou \tCC'$$
  $$ ^t(\neg\CC) = \neg\tCC$$
\end{proposition}
\begin{proof}
  For the conjunction: $ ^t(\CC \et \CC')(O) = (\CC \et \CC')(f(O))= \CC(f(O)) \et \CC'(f(O))=(\tCC \et \tCC')(O)$.
  The proof is similar for the disjunction and the negation.
\end{proof}

Many algorithms deal with conjunctions of anti-monotonic and monotonic constraints.  The two last
propositions mean that these algorithms can be used with the transposed constraints since the
transposed constraint of the conjunction of a monotonic and an anti-monotonic constraint is the
conjunction of a monotonic and an anti-monotonic constraint! The last proposition also helps in
building the transposition of a composed constraint. It is useful for the
algebraisation~\cite{djl02} of the constraint mining problem, where constraints are decomposed in
disjunctions and conjunctions of elementary constraints.

\subsection{Transposed Constraints of Some Classical Constraints}
\label{sec:classical}

In the previous section, we gave the definition of the transposed constraint. In this
definition ($\tCC(O)=\CC(f(O))$), in order to test the transposed constraint on an object set $O$, it is
necessary to compute $f(O)$ (to come back in the attribute space) and then to test $\CC$. This means
that a mining algorithm using this constraint must maintain a dual context, i.e., it must maintain
for each object set~$O$ the corresponding attribute set $f(O)$. Some algorithms already do this, for
instance algorithms which use the so called {\em vertical representation} of the database (like
CHARM~\cite{zaki02}). For some classical constraints however, the transposed constraint can be
rewritten in order to avoid the use of $f(O)$. In this section, we review several classical
constraints and try to find a simple expression of their transposed constraint in the object space.

Let us first consider the minimum frequency constraint: the transposed constraint of
$\Cfreq(A)=(\freq(A) > \gamma)$ is, by definition~\ref{def:tc}, $^t\Cfreq(O)= (\freq(f(O)) >
\gamma)$. By definition of frequency, $\freq(f(O))=\size{g(f(O))}=\size{\cl(O)}$ and if $O$ is a closed
object set, $\cl(O)=O$ and therefore $^t\Cfreq(O)= (\size O > \gamma)$.  Finally, the transposed
constraint of the minimum frequency constraint is the ``minimum size'' constraint.  The
CARPENTER~\cite{pct03} algorithm uses this property and mines the closed patterns in a
divide-and-conquer strategy, stopping when the length of the object set drops below the threshold.

The next two propositions give the transposed constraints of two other classical constraints : 
the subset and superset constraints:

\begin{proposition}[subset constraint transposition]
  \label{prop:sub_ct}
  Let $\CC_{\sseq E}$ be the constraint defined by:
  $\CC_{\sseq E}(A) = (A \sseq E) $ where $E$ is a constant itemset.
  Then if $E$ is closed ($O$ is an object set):
  $$ \tCC_{\sseq E}(O) \equ  g(E) \sseq \cl(O)$$
  and if $E$ is not closed
  $$ \tCC_{\sseq E}(O) \Rightarrow g(E) \sseq \cl(O).$$
\end{proposition}
\begin{proof}
$\tCC_{\sseq E}(O) \equ \CC_{\sseq E}(f(O)) \equ (f(O) \sseq E) \Rightarrow (g(E) \sseq
g(f(O))) \equ (g(E) \sseq \cl(O))$.  Conversely (if $E$ is closed): $(g(E) \sseq g(f(O)))
\Rightarrow (f(O) \sseq \cl(E)) \Rightarrow (f(O) \sseq E)$.
\end{proof}

\begin{proposition}[superset constraint transposition]
  \label{prop:super_ct}
  Let $\CC_{\supseteq E}$ be the constraint defined by:
  $\CC_{\supseteq E}(A) = (A \supseteq E) $ where $E$ is a constant itemset.
  Then:
  $$ \tCC_{\supseteq E}(O) \equ g(E) \supseteq \cl(O).$$
\end{proposition}
\begin{proof}
$\tCC(O) \equ (E \sseq f(O)) \Rightarrow (g(f(O)) \sseq g(E)) \equ (\cl(O)
\sseq g(E))$.  Conversely, $(g(f(O)) \sseq g(E) \Rightarrow (fg(E) \sseq fgf(O)) \Rightarrow fg(E)
\sseq f(O) \Rightarrow \cl(E) \sseq f(O) \Rightarrow E \sseq f(O).$ 
\end{proof}

These two syntactical constraints are interesting because they can be used to construct many other
kind of constraints. In fact, all syntactical constraints can be build on top of these using
conjunctions, disjunctions and negations. With the proposition~\ref{prop:bool_op}, it is then possible to
compute the transposition of many syntactical constraints. Besides, these constraints have
been identified in~\cite{GB00,BGMD032} to formalize dataset reduction techniques.

\begin{table}[htbp]
  \caption{Transposed constraints of some classical constraints. $A$ is a variable closed itemset, 
    $E=\set{e_1,e_2,...,e_n}$ a constant itemset, $O$ a variable closed object set and
    $\overline E = \AA \setminus E=\set{f_1,f_2,...,f_m}$}
  \centering
  \begin{tabular}{cc}
    \ Itemset constraint $\CC(A)$\  & \  Transposed constraint $\tCC(O)$ \ \\\hline
    $\freq(A) \comp \alpha$ & $\size O \comp \alpha$\\
    $A \sseq E$         & if $E$ is closed: $g(E) \sseq O$\\
                        & else: $O \not\sseq g(f_1) \et ... \et O \not\sseq g(f_m) $\\
    $E \sseq A$         & $O \sseq g(E)$\\
    $A \not\sseq E$     & if $E$ is closed: $g(E) \not \sseq O$\\
                        & else: $O \sseq g(f_1) \ou ... \ou O \sseq g(f_m) $\\
    $E \not\sseq A$     &  $O\not \sseq g(E)$\\
    $A\cap E = \emptyset$ & if $\overline E$ is closed:  $ g(\overline E) \sseq O$\\
                          & else: $O \not\sseq g(e_1) \et ... \et O \not\sseq g(e_n) $\\
    $A\cap E \neq \emptyset$ & if $\overline E$ is closed: $ g(\overline E) \not\sseq O$\\
                          & else: $O \sseq g(e_1) \ou ... \ou O \sseq g(e_n) $\\
    $\SUM(A) \comp \alpha$    & $\freq_p(O) \comp \alpha  $\\
    $\MIN(A) \comp \alpha$    &  see text\\
    $\MAX(A) \comp \alpha$    &  see text\\\hline
  \end{tabular}\\
  \smallskip
  $\comp \in \set{<,>,\leq,\geq}$\\
  \label{tab:trans_const}
\end{table}

Table~\ref{tab:trans_const} gives the transposed constraints of several classical constraints if the
object set $O$ is closed (this is not an important restriction since we will use only closed itemsets
extraction algorithms).  These transposed constraints are easily obtained using the two previous
propositions on the superset and the subset constraints and Prop.~\ref{prop:bool_op}. For
instance, if $\CC(A)=(A\cap E \neq \emptyset)$, this can be rewritten $A \not \sseq \overline E$
($\overline E$ denotes the complement of $E$, i.e. $\AA \setminus E$) and then $\neg (A \sseq
\overline E)$. The transposed constraint is therefore, using Prop.~\ref{prop:bool_op} and
\ref{prop:sub_ct}, $\neg (g(\overline E) \sseq O)$ (if $\overline E$ is closed) and finally
$g(\overline E) \not\sseq O$. 
If $\overline E$ is not closed, then we write $E=\set{e_1,...,e_n}$ and we rewrite the constraint 
$\CC(A)=(e_1\in A \ou e_2 \in A \ou ... \ou e_n \in A)$
and then, using  Prop.~\ref{prop:bool_op} and \ref{prop:super_ct}, we obtain the transposed constraint 
 $\tCC(O)=(O\sseq g(e_1) \ou ... \ou O\sseq g(e_n))$.
These expressions are interesting since they do not involve
the computation of $f(O)$. Instead, there are $g(\overline E)$ or $g(e_i)$ ... However, since $E$ is constant,
these values need to be computed only once (during the first database pass, for instance).

\begin{example}
  We show in this example how to compute the transposed constraints with the database of
  Tab.~\ref{tab:database}.  Let the itemset constraint $\CC(A)= (A \cap \at{1}\at{4} \neq
  \emptyset)$.  In the database of Tab.~\ref{tab:database}, the itemset
  $\overline{\at{1}\at{4}}=\at{2}\at{3}$ is closed. Therefore, the transposed constraint is
  (Tab.~\ref{tab:trans_const}) $\tCC(O)=(g(\at{2}\at{3}) \not\sseq O)$.  Since
  $g(\at{2}\at{3})=\ob{1}\ob{2}\ob{3}$, $\tCC(O)=(\ob{1}\ob{2}\ob{3} \not\sseq O)$. The closed object
  sets that satisfy this constraint are $T=\set{\emptyset,\ob{1}\ob{2},\ob{3}}$. If we apply $f$ to
  go back to the itemset space: $\set{f(O) \mid O \in
  T}=\set{\at{1}\at{2}\at{3}\at{4},\at{1}\at{2}\at{3},\at{2}\at{3}\at{4}}$ which are, as expected (and
proved by Th.~\ref{th:1}),
  the closed itemset which satisfy $\CC$.
  
  Consider now the constraint $\CC(A)= (A \cap \at{1}\at{2} \neq \emptyset)$. In this case,
  $\overline{\at{1}\at{2}}=\at{3}\at{4}$ is not closed.  Therefore, we use the second expression in
  Tab.~\ref{tab:trans_const} to compute its transposition.  $\tCC(O) = (O \sseq g(\at{1}) \ou O \sseq
  g(\at{2}))$. Since $g(\at{1})=\ob{1}\ob{2}$ and $g(\at{2})=\ob{1}\ob{2}\ob{3}$, $\tCC(O) = (O \sseq
  \ob{1}\ob{2} \ou O \sseq \ob{1}\ob{2}\ob{3})$ which can be simplified in $\tCC(O) = (O \sseq
  \ob{1}\ob{2}\ob{3})$.  All the closed object sets satisfy this constraint $\tCC$, which is not
  surprising since all the closed itemsets satisfy $\CC$.

  Our last example is the constraint $\CC(A)=(\size{A \cap \at{1}\at{2}\at{4}}\geq 2)$. It can be
  rewritten $\CC(A)=((\at{1}\at{2} \sseq A) \ou (\at{1}\at{4} \sseq A) \ou (\at{2}\at{4} \sseq
  A))$. Using Prop.~\ref{prop:bool_op} and Tab.~\ref{tab:trans_const} we get $\tCC(O)=((O \sseq
  g(\at{1}\at{2})) \ou (O \sseq g(\at{1}\at{4})) \ou (O \sseq g(\at{2}\at{4})))$ which is $\tCC(O)=((O
  \sseq \ob{1}\ob{2}) \ou (O \sseq \emptyset) \ou (O \sseq \ob{3}))$. The closed object sets
  satisfying $\tCC$ are $T=\set{\emptyset, \ob{1}\ob{2},\ob{3}}$ and $\set{f(O) \mid O \in
  T}=\set{\at{1}\at{2}\at{3}\at{4},\at{1}\at{2}\at{3},\at{2}\at{3}\at{4}}$.
\end{example}

Other interesting constraints include aggregate constraints~\cite{ngetal98}. If a numerical value
$a.v$ is associated to each attribute $a\in \AA$, we can define constraints of the form $\SUM(A)
\comp \alpha$ for several aggregate operators such as $\SUM$, $\MIN$, $\MAX$ or $\AVG$, where $\comp \in
\set{<,>,\leq,\geq}$ and $\alpha$ is a numerical value. In this case, $\SUM(A)$ denotes the sum of all
$a.v$ for all attributes $a$ in $A$. 

The constraints $\MIN(A) \comp \alpha$ and $\MAX(A) \comp \alpha$ are special cases of the
constraints of Tab.~\ref{tab:trans_const}. 
For instance, if $sup_\alpha = \set{a \in \AA \mid a.v > \alpha}$ then $\MIN(A)
> \alpha$ is exactly $A \sseq sup_\alpha$ and $\MIN(A) \leq \alpha$ is $A \not\sseq sup_\alpha$.
The same kind of relation holds for $\MAX$ operator: $\MAX(A) > \alpha$ is equivalent to
$A \cap sup_\alpha \neq\emptyset$ and $\MAX(A) \leq \alpha$ is equivalent to 
$A \cap sup_\alpha =\emptyset $. In this case, since $\alpha$ is a constant, the set $sup_\alpha$
can be pre-computed.

The constraints $\AVG(A) \comp \alpha$ and $\SUM(A) \comp \alpha$ are more difficult. Indeed, we only
found one expression (without $f(O)$) for the transposition of $\SUM(A) \comp \alpha$. Its 
transposition is $\tCC(O)=(\SUM(f(O)) \comp \alpha)$. In the database, $f(O)$ is 
a set of attributes, so in the transposed database, it is a set of rows and $O$ is a set
of columns. The values $a.v$ are attached to rows of the transposed database, and $\SUM(f(O))$ is
the sum of these values for the rows containing $O$. Therefore, $\SUM(f(O))$ is a pondered 
frequency of $O$ (in the transposed database) where each row $a$, containing $O$, contributes for $a.v$ in the total (we denote
this pondered frequency by $\freq_p(O)$). It is easy to adapt classical algorithms to count this
pondered frequency. Its computation is the same as the classical frequency except that each
row containing the counted itemset does contribute with a value different from 1 to
the frequency. 

\section{Closed Itemsets Mining}
\label{sec:closed-it}

In a previous work~\cite{rbc03} we showed the complementarity of the set of concepts mined in the
database, with constraining the attribute patterns, and the set of concepts mined in the transposed
database with the {\em negation} of the transposed constraint, when the original constraint is
anti-monotonic. The transposed constraint had to be negated in order to ensure the anti-monotonicity
of the constraint used by the algorithm. This is important because we can keep usual mining
algorithms which deal with anti-monotonic constraint and apply them in the transposed database 
with the negation of the transposed
constraint. We also showed~\cite{rc04} a specific way of mining under monotonic constraint, by
simply mining the transposed database with the transposed constraint (which is anti-monotonic). 
In this section, we generalize these results for more general constraints.

We define the \defnt{constrained closed itemset mining problem}: 
Given a database $\db$ and a constraint $\CC$,
we want to extract all the  closed itemsets
(and their frequencies) satisfying the constraint $\CC$ in the database $\db$. 
More formally, we want to compute the collection:
 $$ \set{(A,\freq(A,\db)) \mid \CC(A,\db) \et \Cclos(A,\db) }.$$ 

The next theorem shows how to compute the above solution set
using the closed object patterns extracted in the transposed database, with the help of the
transposed constraint.

\begin{theorem}
\label{th:1}
  $$ \set{A \mid \CC(A) \et \Cclos(A)}=\set{f(O) \mid \tCC(O) \et \Cclos(O)}.$$
\end{theorem}
\begin{proof}
  By def. \ref{def:tc}, $\set{f(O) \mid \tCC(O) \et \Cclos(O)} = \set{f(O) \mid \CC(f(O)) \et
  \Cclos(O)} = \{A \mid \exists O~s.t.~\CC(A) \et A = f(O) \} = \set{A \mid \CC(A) \et
  \Cclos(A)}$.
\end{proof}

This theorem means that if we extract the collection of all closed object patterns satisfying $\tCC$
in the transposed database, then we can get all the closed patterns satisfying $\CC$ by computing
$f(O)$ for all the closed object patterns. The fact that we only need the {\em closed} object patterns and
not all the object patterns is very interesting since the closed patterns are less numerous and can be
extracted more efficiently (see CHARM~\cite{zaki02}, CARPENTER~\cite{pct03}, CLOSET\cite{peietal00b}
or \cite{boulicautjeudy01a}).  The strategy, which we propose for computing the solution of the
constraint closed itemset mining problem, is therefore:

\begin{enumerate}
\item Compute the transposed constraint $\tCC$ using Tab.~\ref{tab:trans_const} and 
  Prop.~\ref{prop:bool_op}. This step can involve the computation of some constant 
  object sets $g(E)$ used in the transposed constraint.
\item Use one of the known algorithms to extract the constrained closed sets of the transposed database.
  Most closed set extraction algorithms do not use constraints (like CLOSE, CLOSET or 
  CARPENTER). However, it is not difficult to integrate them (by adding more pruning steps)
  for monotonic or anti-monotonic constraints. In~\cite{boulicautjeudy01a}, another algorithm
  to extract constrained closed sets is presented.
\item Compute $f(O)$ for each extracted closed object pattern. In fact, every algorithm already
  computes this when counting the frequency\footnote{This is the frequency in the \em transposed
    database} of $O$, which is $\size{f(O)}$. The frequency of $f(O)$ (in the original database) is
  simply the size of $O$ and can therefore be provided without any access to the database.
\end{enumerate}

The first and third steps can indeed be integrated in the core of the mining algorithm,
as it is done in the CARPENTER algorithm (but only with the frequency constraint).

Finally, this strategy shows how to perform constrained closed itemset mining by
processing all the computations in the transposed database, and using classical algorithms.

\section{Itemsets Mining}
\label{sec:itemset-mining}

In this section, we study how to extract {\em all} the itemsets that satisfy a user constraint (and
not only the closed ones).  We define the \defnt{constrained itemset mining problem} : Given a
database $\db$ and a constraint $\CC$, we want to extract all the itemsets (and their frequencies)
satisfying the constraint $\CC$ in the database $\db$.  More formally, we want to compute the
collection:
$$
\set{(A,\freq(A,\db)) \mid \CC(A,\db) }.$$

In the previous section, we gave a strategy to compute the {\em closed} itemsets satisfying a
constraint. We will of course make use of this strategy. Solving the constrained itemset mining
problem will involve three steps : Given a database $\db$ and a constraint $\CC$,
\begin{enumerate}
\item find a constraint $\CC'$,
\item compute the collection $\set{(A,\freq(A,\db)) \mid \CC'(A,\db) \et \Cclos(A,\db) }$ of closed
  sets satisfying $\CC'$ using the strategy of Sec.~\ref{sec:closed-it},
\item compute the collection $\set{(A,\freq(A,\db)) \mid \CC(A,\db) }$ of all the itemsets
  satisfying $\CC$ from the closed ones satisfying $\CC'$.
\end{enumerate}

We will study the first step in the next subsection and the third one in
Sec.~\ref{sec:regeneration}, but first we will show why it is necessary to introduce a new
constraint $\CC'$. Indeed, it is not always possible to compute all the itemsets that satisfy $\CC$
from the closed sets that satisfy $\CC$. Let us first recall how the third step is done in the
classical case where $\CC$ is the frequency constraint~\cite{pasquieretal99}:

The main used property is that all the itemsets of an equivalence class have the same frequency than
the closed itemset of the class. Therefore, if we know the frequency of the closed itemsets, it is
possible to deduce the frequency of non-closed itemsets provided we are able to know in which class
they belong. The regeneration algorithm of~\cite{pasquieretal99} use a top down approach. Starting from the
largest frequent closed itemsets, it generates their subsets and assign them their frequencies, until
all the itemsets have been generated.

\begin{figure}[htbp]
  \centering \includegraphics[scale = 1]{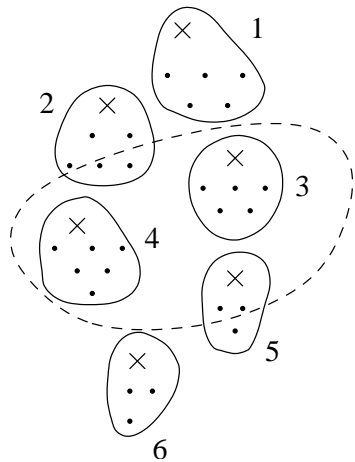}
  \caption{The dots represent itemsets, the x are closed itemsets, the lines enclose the equivalence
    classes. The itemsets inside the region delimited by the dashed line satisfy the constraint
    $\CC$ and the others do not. The closed sets satisfying $\CC$ are the closed sets of classes 3,
    4 and 5. They will enable to generate the itemsets of these three classes. However, to get the
    two itemsets of class 2, we need the closed itemset of this class which does not satisfy $\CC$.
    Therefore, in this case, having the closed itemsets satisfying $\CC$ is not enough to generate
    all itemsets satisfying $\CC$.}
  \label{fig:regen_pb}
\end{figure}

Now, assume that the constraint $\CC$ is not the frequency constraint and that we have computed all
the closed itemsets (and their frequencies) that satisfy $\CC$. If an itemset satisfies $\CC$, it is
possible that its closure does not satisfies it. In this case, it is not possible to compute the
frequency of this itemset from the collection of the closed itemsets that satisfy $\CC$ (this is
illustrated in Fig.~\ref{fig:regen_pb}). Finally, the collection of the closed itemsets satisfying
$\CC$ is not sufficient to generate the non-closed itemsets. In the next section, we show how the
constraint $\CC$ can be relaxed to enable the generation all the non-closed itemsets satisfying it.

\subsection{Relaxation of the Constraint}
\label{sec:relax-constr}
In order to be able to generate all the itemsets from the closed ones, it is necessary to have at least the
collection of closed itemsets of all the equivalence classes that contain an itemset satisfying the
constraint $\CC$. This collection is also the collection of the closures of all itemsets satisfying
$\CC$ : $\set{\cl(A) \mid \CC(A,\db) }$.

\begin{figure}[htbp]
  \centering \includegraphics[scale = 1]{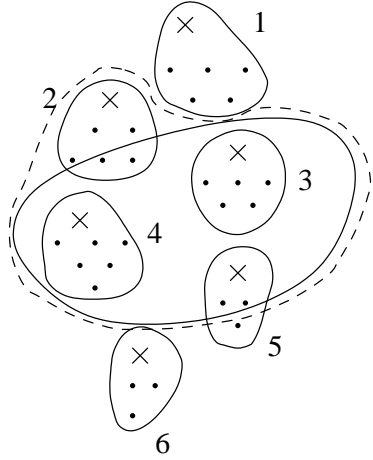}
  \caption{An optimal relaxation of $\CC$. The constraint $\CC$ is represented by the solid line
    and an optimal relaxation is represented by the dashed line.}
  \label{fig:relax}
\end{figure}

We must therefore find a constraint $\CC'$ such that $\set{\cl(A) \mid \CC(A,\db)}$ is included in 
$\set{A \mid \CC'(A,\db) \et \Cclos(A) } $. We call such a $\CC'$ constraint a \defnt{good relaxation} of $\CC$
(see Fig.~\ref{fig:relax}). If we have an equality instead of the inclusion, we call $\CC'$ an
\defnt{optimal relaxation} of $\CC$. For example, the constant ``true'' constraint (which is true on all itemset) is a good
relaxation of any constraint, however it is not very interesting since it will not provide any
pruning opportunity during the extraction of step~2.

If the closed itemsets (and their frequencies) satisfying an optimal relaxation of $\CC$ are computed
in step~2, we will have enough information for regenerating all itemsets satisfying $\CC$ in
step~3.  However it is not always possible to find such an optimal relaxation. In this case, we can
still use a good relaxation in step~2. In this case, some superfluous closed itemsets will be
present in the collection and will have to be filtered out in step~3.

\vspace*{0.3cm}

We will now give optimal relaxation for some classical constraints, and we start with two trivial
cases :

\begin{proposition}
  \label{prop:monot}
  The optimal relaxation of a monotonic constraint is the constraint itself and the optimal
  relaxation of the frequency constraint is the frequency constraint itself.
\end{proposition}

\begin{proof}
  Let $\CC$ be a monotonic constraint or a frequency constraint.  We only have to prove that if an
  itemset $A$ satisfy $\CC$ then $\cl(A)$ also.  If $\CC$ is monotonic, this is true since $S\sseq
  \cl(S)$ (Prop.~\ref{prop:decr}.  If $\CC$ is a minimum frequency constraint, it is true because $A$
  and $\cl(A)$ have the same frequency.
\end{proof}

The next proposition is used to compute the relaxation of a complex constraint from the relaxation
of simple constraints.

\begin{proposition}
  \label{prop:etou}
  If $\CC_1$ and $\CC_2$ are two constraints and $\CC'_1$ and $\CC'_2$ are optimal relaxation of
  them, then :
  \begin{itemize}
  \item $\CC'_1 \ou \CC'_2$ is an optimal relaxation of $\CC_1 \ou \CC_2$ and
  \item $\CC'_1 \et \CC'_2$ is a good relaxation of $\CC_1 \et \CC_2$.
  \end{itemize}
\end{proposition}

\begin{proof}
  A constraint $\CC'$ is a good relaxation of a constraint $\CC$ if and only if $\forall A, \CC(A)
  \Rightarrow \CC'(\cl(A))$. To prove that it is an {\em optimal} relaxation, we must also prove
  that if $A$ is closed and satisfies $\CC'$ then there exists an itemset $B$ satisfying $\CC$ such
  that $\cl(B)=A$ (cf. definitions). We will use this two facts in our proofs.\\
  Let $A$ be an itemset satisfying $\CC_1 \et \CC_2$. This means that $A$ satisfies $\CC_1$ and
  $\CC_2$. Therefore, $\cl(A)$ satisfies $\CC'_1$ and $\CC'_2$, i.e., $cl(A)$ satisfies $\CC'_1 \et
  \CC'_2$.This means that $\CC'_1 \et \CC'_2$ is a good relaxation of $\CC_1 \et \CC_2$.\\
  We can prove similarly that $\CC'_1 \ou \CC'_2$ is a good relaxation of $\CC_1 \ou \CC_2$.  Let us
  now prove that it is optimal: Let $A$ be a closed itemset satisfying $\CC'_1 \ou \CC'_2$.  Then
  $A$ satisfies $\CC'_1$ or $\CC'_2$, suppose that it satisfies $\CC'_1$. Since $\CC'_1$ is an
  optimal relaxation of $\CC_1$, there exists $B$ satisfying $\CC_1$ such that $\cl(B)=A$.
  Therefore $B$ satisfies $\CC_1 \ou \CC_2$ and $\cl(B)=A$.
\end{proof}

We found no relaxation for the negation of a constraint but this is not a problem. If the constraint
is simple (i.e., in Tab.~\ref{tab:trans_const}) its negation is also in the table and if it is
complex, then we can ``push'' the negation into the constraint as shown in the next example.

\begin{example}
  \label{ex:neg}
  Let $\CC(A)=(\neg (((\freq(A)>3) \et (A \not\sseq E)) \ou (A \cap F = \emptyset)))$ where $E$ and
  $F$ are two constant itemsets.  We can push the negation and we get: $\CC(A)= ((\neg (\freq(A)>3)
  \ou \neg(A \not\sseq E)) \et \neg (A \cap F = \emptyset))$, and finally :
  $$\CC(A)= (((\freq(A) \leq 3) \ou (A \sseq E)) \et (A \cap F \neq \emptyset)).$$
  Then with
  Prop.~\ref{prop:monot},~\ref{prop:etou} and Tab.~\ref{tab:rel}, we can compute a good relaxation
  $\CC'$ of $\CC$:
  $$\CC'(A)= (((\freq(A) \leq 3) \ou (A \sseq \cl(E))) \et (A \cap F \neq \emptyset)).$$
\end{example}

Table~\ref{tab:rel} gives good relaxation of the other constraints of Tab.~\ref{tab:trans_const}
which are not covered by the previous proposition (i.e., which are not monotonic) except for the
non-monotonic constraints involving $\SUM$ for which we did not find any interesting (i.e., other
than the constant true constraint) good relaxation.

\begin{table}[htbp]
  \caption{Good relaxation of some classical constraints. $A$ is a variable closed itemset, 
    $E=\set{e_1,e_2,...,e_n}$ a constant itemset.}
  \centering
  \begin{tabular}{cc}
    \ Itemset constraint $\CC(A)$\  & \  Good relaxation $\CC'(A)$\ \\\hline
    $A \sseq E$         &  $A \sseq \cl(E)$ \\
    $E \not\sseq A$     &  $A \sseq \cl(\overline{e_1}) \ou A \sseq \cl(\overline{e_2}) \ou ... \ou 
    A \sseq \cl(\overline{e_n})$\\
    $A\cap E = \emptyset$ &  $A \sseq \cl(\overline E)$ \\
    $\MIN(A) > \alpha$    &  $A \sseq \cl(sup_\alpha)$\\
    $\MAX(A) < \alpha$    &  $A \sseq  \cl(\overline{supeq_\alpha})$ \\\hline
  \end{tabular}
  \label{tab:rel}
\end{table}

\begin{proof}
  We prove here the results given in Tab.~\ref{tab:rel}.\\
  $\CC(A)=(A \sseq E)$, $\CC'(A)=(A \sseq \cl(E))$: If $A \sseq E$ then $\cl(A) \sseq \cl(E)$.
  This means that $\CC(A) \Rightarrow \CC'(\cl(A))$ therefore $\CC'$ is a good relaxation of $\CC$.\\
  
  $\CC(A)=(A\cap E = \emptyset)$: $\CC$ can be rewritten $\CC(A)= (A\sseq \overline E)$ and the
  previous case applies with $\overline E$ instead of $E$.\\
  
  $\CC(A)=(E \not\sseq A)$: If $E=\set{e_1,e_2,...,e_n}$, this constraint can be rewritten
  $\set{e_1} \not \sseq A \ou \set{e_2} \not \sseq A \ou \ldots \ou \set{e_n} \not \sseq A$ which is
  also $A \sseq \overline{\set{e_1}} \ou \ldots \ou A \sseq \overline{\set{e_n}}$.
  Then the first case and Prop~\ref{prop:etou} give the result.\\
  
  $\CC(A)=(\MIN(A) > \alpha)$ and $\CC(A)=(\MAX(A) < \alpha)$: $\CC(A)=(\MIN(A) > \alpha)$ can be
  rewritten $A \sseq sup_\alpha$ with $sup_\alpha = \set{a \in \AA \mid a.v > \alpha}$ and we are in
  the first case.  $\CC(A)=(\MAX(A) < \alpha)$ can be rewritten $A \cap supeq_\alpha =\emptyset$
  with $supeq_\alpha = \set{a \in \AA \mid a.v \geq \alpha}$ and we are in the second case.
\end{proof}

\subsection{Regeneration}
\label{sec:regeneration}

Given a database $\db$ and a constraint $\CC$, we suppose in this section that a collection
$\set{(A,\freq(A,\db)) \mid \CC'(A,\db) \et \Cclos(A,\db)} $ of closed itemsets (and their
frequencies) satisfying a good relaxation $\CC'$ of $\CC$ is available.  The aim is to compute the
collection $\set{(A,\freq(A,\db)) \mid \CC(A,\db)} $ of all itemset satisfying $\CC$ (and their
frequencies).

If $\CC$ is a minimum frequency constraint, $\CC$ is an optimal relaxation of itself, therefore
we take $\CC'=\CC$. The regeneration algorithm is then the classical algorithm~6 of~\cite{pasquieretal99}.
We briefly recall this algorithm:

We suppose that the frequent closed itemsets (and their frequencies) 
of size $i$ are stored in the list $\LL_i$ for
$0<i\leq k$ where $k$ is the size of the longest frequent closed itemset.
At the end of the algorithm, each $\LL_i$ contains all the frequent itemsets of size $i$ and their
frequencies.\\

\noindent1 for $(i=k;i>0;i--)$\\
2 \ \ forall $A \in \LL_i$\\
3 \ \ \ \ forall subset $B$ of size $(i-1)$ of $A$\\
4 \ \ \ \ \ \ if $B \not\in \LL_{i-1}$\\
5 \ \ \ \ \ \ \ \ $B.freq = A.freq$\\
6 \ \ \ \ \ \ \ \ $\LL_{i-1} = \LL_{i-1} \cup \set{B}$\\
7 \ \ \ \ \ \ endif\\
8 \ \ \ \ end\\
9 \ \ end\\
10 end\\

If $\CC'$ is not the frequency constraint, this algorithm generates all the subsets of 
the closed itemsets satisfying $\CC'$ and two problems arise:
\begin{enumerate}
\item Some of these itemsets do not satisfy $\CC$. For instance, in Fig.~\ref{fig:relax}, all the
itemsets of classes 2, 3, 4, 5 and 6 are generated (because they are subsets of closed itemsets that
satisfy $\CC'$) and only those of classes 3 and 4 and some of classes 2 and 5 satisfy $\CC$.
\item The frequency computed in step 5 of the above algorithm for $B$ is correct only if the closure 
  of $B$ is in the collection of the closed sets at the beginning of the algorithm. 
  If it is not, then this computed frequency is smaller
  than the true frequency of $B$. In Fig.~\ref{fig:relax}, this means that the computed frequency
  of the itemsets of class 6 are not correct. 
\end{enumerate}

However, the good news is that all the itemsets satisfying $\CC$ are generated (because $\CC'$ is a 
good relaxation of $\CC$) and their computed frequencies are correct (because their closures belongs
to the $\LL_i$ at the beginning). 

A last filtering phase is therefore necessary to filter out all the generated itemsets that 
do not satisfy $\CC$. This phase can be pushed inside the above generation algorithm 
if the constraint $\CC$ has good properties (particularly if it is a conjunction of a monotonic
part and an anti-monotonic one). However, we will not detail this point here.

\vspace*{0.3cm}

We are still facing a last problem: to test $\CC(A)$, we can need $\freq(A)$. However,
if $\CC(A)$ is false, it is possible that the computed frequency of $A$ is not correct. To solve
this problem, we propose the following strategy.

We assume that the constraint $\CC$ is a Boolean formula built using
the atomic constraints listed
in Tab.~\ref{tab:trans_const} and using the two operators $\et$ and $\ou$ (if the $\neg$
operator appears, it can be pushed inside the formula as shown in Ex.~\ref{ex:neg}).
Then, we rewrite this constraint in disjunctive normal form (DNF), i.e., 
$\CC= \CC_1 \ou \CC_2 \ou \ldots \ou \CC_n$ with $\CC_i=\AA_{m_{i-1}+1} \et \ldots \et \AA_{m_i}$
where each $\AA_i$ is a constraint listed in Tab.~\ref{tab:trans_const}.

Now, consider an itemset $A$ whose computed frequency is $f$ (with $f \leq \freq(A)$).
First, we consider all the conjunction $\CC_i$ that we can compute, this include those 
where $\freq(A)$ does not appear and those of the form $\freq(A) > \alpha$ or $\freq(A) < \alpha$ 
where $\alpha < f$ (in this two cases we can conclude since $\freq(A) \geq f$).
If one of them is true, then $\CC(A)$ is true and $A$ is not filtered out. 

If all of them are
false, we have to consider the remaining conjunctions of the form
$\AA_1 \et \ldots \et (\freq(A) > \alpha) \et \ldots$ with $\alpha \geq f$.
If one of the $\AA_i$ is false,
then the conjunction is false. If all are true, we suppose that $\freq(A) > \alpha$:
in this case $\CC(A)$ is true and therefore $\freq(A)=f$ which contradict $\alpha \geq f$.
Therefore, $\freq(A) > \alpha$ is false and also the whole conjunction.

If it is still impossible to answer, it means that all the conjunctions are false, and that 
there are conjunction of the form  
$\AA_1 \et \ldots \et (\freq(A) < \alpha) \et \ldots$ with $\alpha \geq f$. In this case, 
it is not possible to know if $\CC(A)$ is true without computing the frequency $\freq(A)$.

Finally, all this means that if there is no constraints of the form $\freq(A) < \alpha$ in 
the DNF of $\CC$, we can do this last filtering phase efficiently. If it appears, then
the filtering phase can involve access to the database to compute the frequency of some
itemsets. Of course, all these frequency 
computation should be made in one access to the database.

\begin{example}
In this example, we illustrate the complete process of the resolution of the constrained itemset 
mining problem on two constraints (we still use the dataset of Tab.~\ref{tab:database}):\\

$\CC(A)=((\freq(A)>1) \ou (\at 1 \in A))$.\\
This constraint is its own optimal relaxation (cf. Prop.~\ref{prop:monot} and~\ref{prop:etou}).
According to Tab.~\ref{tab:trans_const} and Prop.~\ref{prop:bool_op}, its transposed
constraint is $\tCC(O)= ((\size O > 1) \ou (O \sseq g(\at 1)))$ and $g(\at 1)=\ob 1\ob 2$.
The closed objects sets that satisfy this constraints are 
$T= \set{\ob 1 \ob 2,\ob 1 \ob 2 \ob 3,\emptyset }$.
 If we apply $f$ to
go back to the itemset space: 
$\set{f(O) \mid O \in T}=\set{\at{1}\at{2}\at{3}\at{4},\at{1}\at{2}\at{3},\at{2}\at{3}}$.
Since this set contains $\at{1}\at{2}\at{3}\at{4}$, all the itemsets are generated. However, the
generated frequency for the itemsets of the class of $\at{2}\at{3}\at{4}$ is 0. The other generated
frequencies are correct.
$\CC$ is in DNF with two simple constraints ($\freq(A) > 1)$ and $(\at 1 \in A)$.
During the filtering step, when considering the itemsets of $\at{2}\at{3}\at{4}$'s class,
 the second constraint is always true. Since the generated frequency $f$ is 0 and $\alpha$
is 1, $\alpha>f$ and therefore these itemsets must be filtered out. Finally, the remaining 
itemsets are exactly those that satisfy $\CC$.\\

$\CC(A)=((\freq(A)>1) \et (A \sseq \at 2\at 4))$.\\
A good relaxation of $\CC$ is $\CC'(A)=((\freq(A)>1) \et (A \sseq \cl(\at 2\at 4))) =
((\freq(A)>1) \et (A \sseq \at 2\at 3\at 4))$. The corresponding transposed constraint is
$\tCC'(O)= ((\size O > 1) \et (g(\at 2\at 3\at 4) \sseq O)) =
((\size O > 1) \et (\ob 3 \sseq O))$ since $\at 2\at 3\at 4$ is closed.
The closed objects sets that satisfy this constraints are 
$T= \set{\ob 1 \ob 2 \ob 3}$.
 If we apply $f$ to
go back to the itemset space: 
$\set{f(O) \mid O \in T}=\set{\at{2}\at{3}}$.
Then all the subsets of $\at{2}\at{3}$ are generated and only $\emptyset$ and $\at 2$ remains
after the filtering step.
\end{example}

\section{Conclusion}
\label{sec:conclusion}

In order to mine constrained closed patterns in databases with more columns than rows, we proposed a
complete framework for the transposition: we gave the expression in the transposed database of the
transposition of many classical constraints, and showed how to use existing closed set mining
algorithms (with few modifications) to mine in the transposed database. 

Then we gave a strategy to use this framework to mine all the itemset satisfying a constraint
when a constrained closed itemset mining algorithm is available. This strategy consists 
of three steps: generation of a relaxation of the constraint, extraction of the closed 
itemset satisfying the relaxed constraint and, finally, generation of all the itemsets 
satisfying the original constraint.

We can therefore choose the smallest space between the object space and the attribute space depending
on the number of rows/columns in the database. Our strategy gives new opportunities for
the optimization of mining queries (also called inductive queries) in contexts having a pathological
size. This transposition principle could also be used for the optimization of sequences of queries:
the closed object sets computed in the transposed database during the evaluation of previous queries
can be stored in a cache and be re-used to speed up evaluation of new queries in a fashion similar
to~\cite{jeudyboulicaut02b}.


\begin{thebibliography}{10}

\bibitem{agrawaletal96}
R.~Agrawal, H.~Mannila, R.~Srikant, H.~Toivonen, and A.~I. Verkamo.
\newblock Fast discovery of association rules.
\newblock In {\em Advances in Knowledge Discovery and Data Mining}, 1996.

\bibitem{bastideetal00b}
Y.~Bastide, N.~Pasquier, R.~Taouil, G.~Stumme, and L.~Lakhal.
\newblock Mining minimal non-redundant association rules using frequent closed
  itemsets.
\newblock In {\em Proc. Computational Logic}, volume 1861 of {\em LNAI}, pages
  972--986, 2000.

\bibitem{brb04}
J.~Besson, C.~Robardet, and J.-F. Boulicaut.
\newblock Constraint-based mining of formal concepts in transactional data.
\newblock In {\em Proc. PAKDD}, 2004.
\newblock to appear.

\bibitem{BGMD032}
F.~Bonchi, F.~Giannotti, A.~Mazzanti, and D.~Pedreschi.
\newblock Exante: Anticipated data reduction in constrained pattern mining.
\newblock In {\em PKDD'03}, 2003.

\bibitem{bgkm02}
E.~Boros, V.~Gurvich, L.~Khachiyan, and K.~Makino.
\newblock On the complexity of generating maximal frequent and minimal
  infrequent sets.
\newblock In {\em Symposium on Theoretical Aspects of Computer Science}, pages
  133--141, 2002.

\bibitem{bbr03}
J.-F. Boulicaut, A.~Bykowski, and C.~Rigotti.
\newblock Free-sets : a condensed representation of boolean data for the
  approximation of frequency queries.
\newblock {\em DMKD}, 7(1), 2003.

\bibitem{boulicautjeudy01a}
J.-F. Boulicaut and B.~Jeudy.
\newblock Mining free-sets under constraints.
\newblock In {\em Proc. IDEAS}, pages 322--329, 2001.

\bibitem{bgk02}
C.~Bucila, J.~Gehrke, D.~Kifer, and W.~White.
\newblock Dualminer: a dual-pruning algorithm for itemsets with constraints.
\newblock In {\em Proc. SIGKDD}, pages 42--51, 2002.

\bibitem{caldersetal02}
T.~Calders and B.~Goethals.
\newblock Mining all non-derivable frequent itemsets.
\newblock In {\em Proc. PKDD}, volume 2431 of {\em LNAI}, pages 74--85, 2002.

\bibitem{deraedtkramer01}
L.~de~Raedt and S.~Kramer.
\newblock The levelwise version space algorithm and its application to
  molecular fragment finding.
\newblock In {\em Proc. IJCAI}, pages 853--862, 2001.

\bibitem{dl99}
G.~Dong and J.~Li.
\newblock Efficient mining of emerging patterns : discovering trends and
  differences.
\newblock In {\em Proc. SIGKDD}, pages 43--52, 1999.

\bibitem{fmn03}
H.~Fu and E.~M. Nguifo.
\newblock How well go lattice algorithms on currently used machine learning
  testbeds ?
\newblock In {\em 1st Intl. Conf. on Formal Concept Analysis}, 2003.

\bibitem{GB00}
B.~Goethals and J.~V. den Bussche.
\newblock On supporting interactive association rule mining.
\newblock In {\em DAWAK'00}, 2000.

\bibitem{jeudyboulicautida}
B.~Jeudy and J.-F. Boulicaut.
\newblock Optimization of association rule mining queries.
\newblock {\em Intelligent Data Analysis}, 6(4):341--357, 2002.

\bibitem{jeudyboulicaut02b}
B.~Jeudy and J.-F. Boulicaut.
\newblock Using condensed representations for interactive association rule
  mining.
\newblock In {\em Proc. PKDD}, volume 2431 of {\em LNAI}, 2002.

\bibitem{ngetal98}
R.~Ng, L.~V. Lakshmanan, J.~Han, and A.~Pang.
\newblock Exploratory mining and pruning optimizations of constrained
  associations rules.
\newblock In {\em SIGMOD}, 1998.

\bibitem{mn00}
E.~M. Nguifo and P.~Njiwoua.
\newblock {GLUE}: a lattice-based constructive induction system.
\newblock {\em Intelligent Data Analysis}, 4(4):1--49, 2000.

\bibitem{pct03}
F.~Pan, G.~Cong, A.~K.~H. Tung, J.~Yang, and M.~J. Zaki.
\newblock {CARPENTER}: Finding closed patterns in long biological datasets.
\newblock In {\em Proc. SIGKDD}, 2003.

\bibitem{pasquieretal99}
N.~Pasquier, Y.~Bastide, R.~Taouil, and L.~Lakhal.
\newblock Efficient mining of association rules using closed itemset lattices.
\newblock {\em Information Systems}, 24(1):25--46, Jan. 1999.

\bibitem{peietal01}
J.~Pei, J.~Han, and L.~V.~S. Lakshmanan.
\newblock Mining frequent itemsets with convertible constraints.
\newblock In {\em Proc. ICDE}, pages 433--442, 2001.

\bibitem{peietal00b}
J.~Pei, J.~Han, and R.~Mao.
\newblock {CLOSET} an efficient algorithm for mining frequent closed itemsets.
\newblock In {\em Proc. DMKD workshop}, 2000.

\bibitem{djl02}
L.~D. Raedt, M.~Jaeger, S.~Lee, and H.~Mannila.
\newblock A theory of inductive query answering (extended abstract).
\newblock In {\em Proc. ICDM}, pages 123--130, 2002.

\bibitem{rbc03}
F.~Rioult, J.-F. Boulicaut, B.~Crémilleux, and J.~Besson.
\newblock Using transposition for pattern discovery from microarray data.
\newblock In {\em DMKD workshop}, 2003.

\bibitem{rc04}
F.~Rioult and B.~Crémilleux.
\newblock Optimisation of pattern mining : a new method founded on database
  transposition.
\newblock In {\em EIS'04}, 2004.

\bibitem{scr042}
A.~Soulet, B.~Crémilleux, and F.~Rioult.
\newblock Condensed representation of emerging patterns.
\newblock In {\em Proc. PAKDD}, 2004.

\bibitem{SS02}
B.~Stadler and P.~Stadler.
\newblock Basic properties of filter convergence spaces.
\newblock {\em J. Chem. Inf. Comput. Sci.}, 42, 2002.

\bibitem{WIL92}
R.~Wille.
\newblock Concept lattices and conceptual knowledge systems.
\newblock In {\em Computer mathematic applied, 23(6-9):493-515}, 1992.

\bibitem{zaki02}
M.~J. Zaki and C.-J. Hsiao.
\newblock {CHARM}: An efficient algorithm for closed itemset mining.
\newblock In {\em Proc. SDM}, 2002.

\end{thebibliography}
\end{document}